\documentclass{article}

\usepackage[final]{isic_2018}

\usepackage[utf8]{inputenc} 
\usepackage[T1]{fontenc}    
\usepackage{hyperref}       
\usepackage{url}            
\usepackage{booktabs}       
\usepackage{amsfonts}       
\usepackage{nicefrac}       
\usepackage{microtype}      
\usepackage{graphicx}
\usepackage{subfigure}

\title{Dermoscopic Image Analysis for ISIC Challenge 2018}

\author{
  Jinyi Zou \quad\quad\quad Xiao Ma \quad\quad\quad Cheng Zhong \\
  \texttt{\{zoujy2,maxiao3,zhongcheng3\}@lenovo.com} \\
  \And
  Yao Zhang \\
  \texttt{zhangyao215@mails.ucas.ac.cn} \\
}

\begin{document}

\maketitle

\begin{abstract}
  This short paper reports the algorithms we used and the evaluation performances for ISIC Challenge 2018. Our team participates in all the tasks in this challenge. In lesion segmentation task, the pyramid scene parsing network (PSPNet) is modified to segment the lesions. In lesion attribute detection task, the modified PSPNet is also adopted in a multi-label way. In disease classification task, the DenseNet-169 is adopted for multi-class classification.
\end{abstract}

\section{Introduction}
The goal of the recurring ISIC challenge is to develop image analysis algorithms to enable the automated diagnosis of melanoma from dermoscopic images. In ISIC 2018, the challenge is broken into three separate tasks: lesion segmentation, lesion attribute detection and disease classification. For task1 and task2, we adopt a modified PSPNet for lesion segmentation [1]. For task3, we adopt the DenseNet-169 model for classification [2].

%
%
%
%
%
%
%
%
%
%
%

\section{Lesion Segmentation}
\label{gen_inst}

The goal of this task is the automated predictions of lesion segmentation boundaries within dermoscopic images. There are 2594 images for training and 100 images for validation. The evaluation metrics in this task is Jaccard Index, as shown in Eq.\ref{jac}. Different with ISIC 2017, the segmentation score below 0.65 will be set to 0 in ISIC 2018.

\begin{equation}
	J(A, B)=\frac{|A \cap B|}{|A \cup B|}, \quad 0 \le J(A, B) \le 1,
\label{jac}
\end{equation}
where $A$ denotes the number of predicted pixels belonging to the lesion and $B$ denotes the number of ground truth pixels belonging to the lesion. In the following, I will describe the data augmentation strategies, the model design and experimental results in lesion segmentation task.

\subsection{Data Augmentation}

The data augmentation strategy can be summarized as follows:
\begin{itemize}
    \item Image flip in both horizontal and vertical directions
    \item Image scale variation from 0.8 to 1.2
    \item Color jitter using the parameters the brightness with a maximum delta of 64/255, contrast with a maximum delta of 0.75, saturation with a maximum delta of 0.25 and hue with a maximum delta of 0.04
\end{itemize}

In our experiments we find that color jitter makes a little contribution for lesion segmentation.

\subsection{Model Design and Experimental Results}

\begin{figure}[!tp]
\centering
\includegraphics[width=\linewidth]{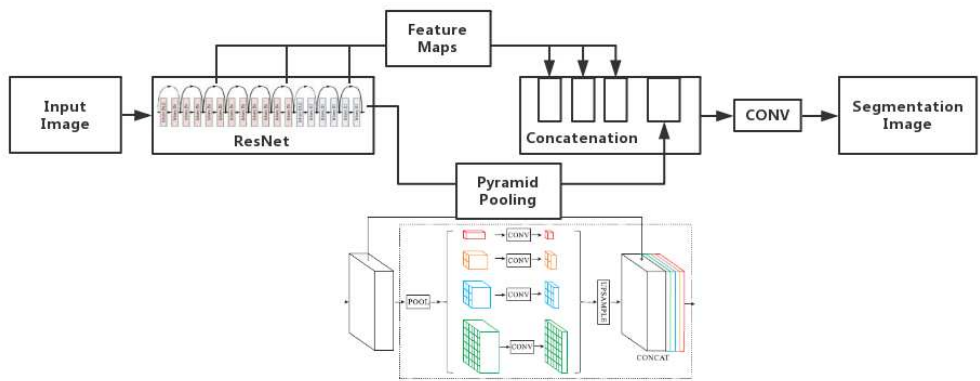}	
\caption{The flowchart of modified PSPNet.}
\label{psp}
\end{figure}

To extract low-level image features to enhance the segmentation performance, we not only use the features maps in last layer but also the feature maps in middle layers in ResNet for concatenation, as shown in Fig.\ref{psp}. Therefore, we can make use of more finer texture information in the modified PSPNet to improve lesion segmentation performance.

In the training stage, the parameters of the model can be summarized as follows: the backbone network is ResNet, the training image resolution is $720 \times 720$, the learning rate is set to $1e-5$ with Adam Optimizer. In addition, the post-processing procedure including the conditional random field (CRF) and the watershed algorithms are adopted to filter the segmentation noises and only the largest connected component will be remained as the lesion segmentation result [3][4].

\begin{figure}[!tp]
\centering
\includegraphics[width=\linewidth]{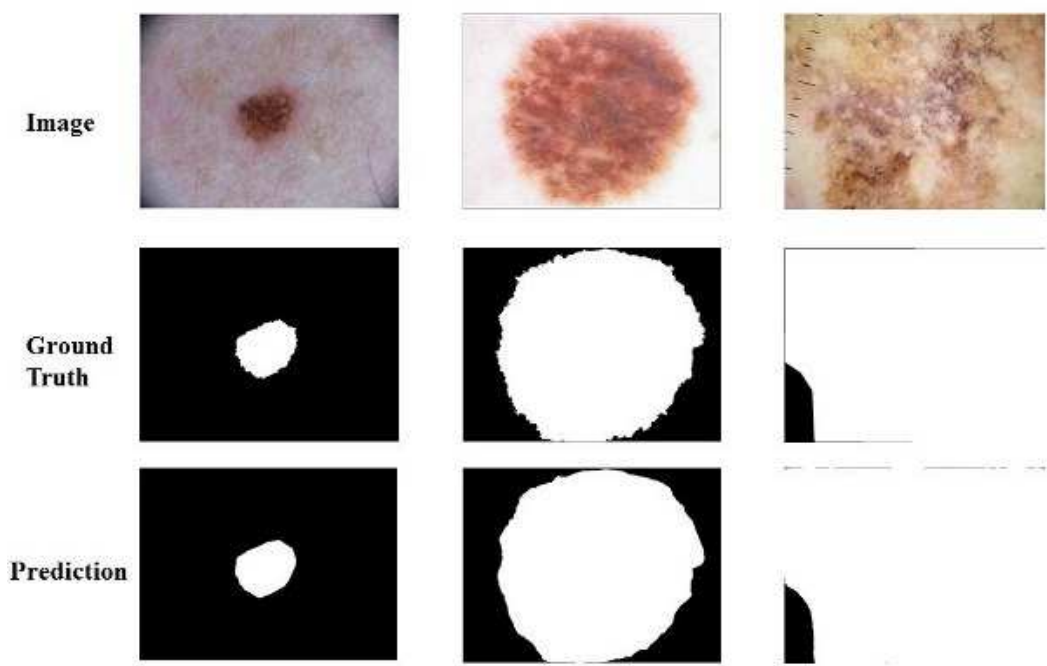}	
\caption{Experimental results for lesion segmentation using PSPNet.}
\label{psp_lesion_seg}
\end{figure}

Some experimental results are shown in Fig.\ref{psp_lesion_seg}. In training dataset, the segmentation performance is $0.957$. In validation dataset, the segmentation performance is $0.775$ without post-processing and $0.780$ with post-processing.

\section{Lesion Attribute Detection}

The goal of this task is the automated predictions of the locations of dermoscopic attributes within dermoscopic images, including pigment network, negative network, streaks, milia-like cysts and globules (including dots). This task shares the same training data with task1, with different mask data for different segmentation targets. The evaluation is similar to task1, the difference is that in this task we need to predict multiple categories.

In this task, the modified PSPNet is also adopted to achieve lesion attribute detection in a multi-class way. In addition, the lesion segmentation result in task1 is used to refine the lesion attribute detection results. Only the pixels in segmented lesions will be categorized to lesion attributes.

\begin{figure}[!tp]
\centering
\includegraphics[width=\linewidth]{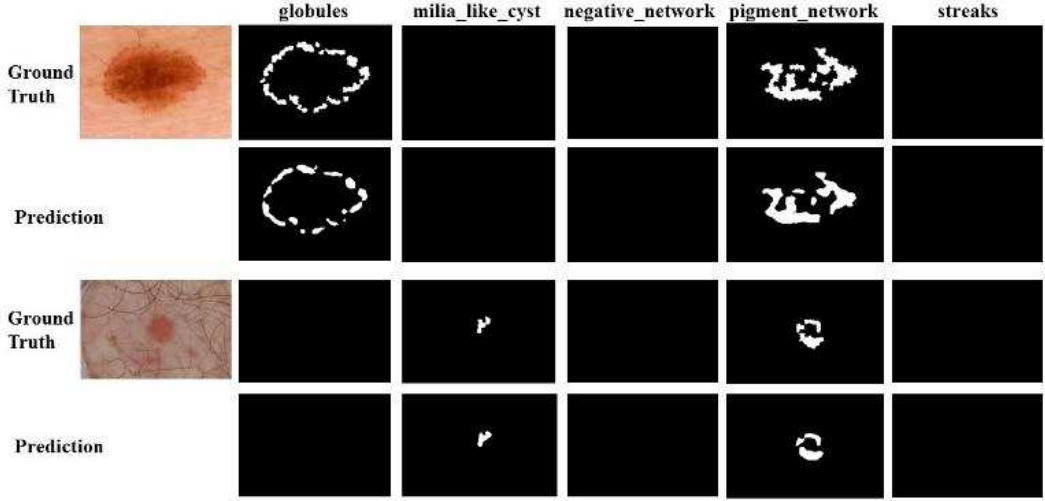}	
\caption{Experimental results for using PSPNet for lesion attribute detection.}
\label{psp_lesion_attri}
\end{figure}

The training parameters for PSPNet in this task are the same with task1. Some experimental results are shown in Fig.\ref{psp_lesion_attri}. In training dataset, the segmentation performances for different types are: $0.695$ for pigment network, $0.617$ for negative network, $0.758$ for streaks, $0.779$ for milia-like cysts and $0.672$ for globules; In validation dataset the segmentation performance is $0.40$.

\section{Disease Classification}

The goal of this task is the automated predictions of disease classification within dermoscopic images, including melanoma, melanocytic nevus, basal cell carcinoma, actinic keratosis / bowen’s disease (intraepithelial carcinoma), benign keratosis (solar lentigo / seborrheic keratosis / lichen planus-like keratosis), dermatofibroma and vascular lesion. There are 10015 images for training and 193 images for validation. The evaluation metrics in this task is the normalized multi-class accuracy (balanced across categories). Compared with the previous challenges, we need to compute the diagnosis probabilities for multiple classes in ISIC 2018, whereas we only need to make a binary decision “biopsy” versus “don’t biopsy” in ISIC 2017. In the following, I will describe the data augmentation strategies, the model design and experimental results in disease classification task.

\subsection{Data Augmentation}

The data augmentation strategy can be summarized as follows:
\begin{itemize}
    \item Image flip in both horizontal and vertical directions
    \item Image scale variation from 0.8 to 1.2
\end{itemize}

Because we used the color constancy here, the color jitter is not used for data augmentation [5].

\subsection{Model Design and Experimental Results}

\begin{figure}
  \centering
  \subfigure[The illustration of hierarchical structure.]{
    \label{hierarchical} 
    \includegraphics[width=2.5in]{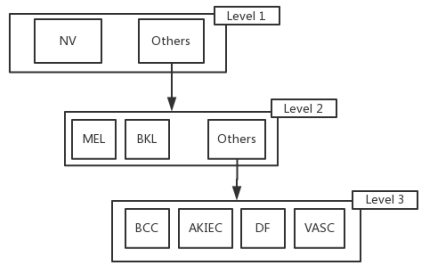}}
  \hspace{0.5in}
  \subfigure[The illustration of DenseNet.]{
    \label{densenet} 
    \includegraphics[width=2.0in]{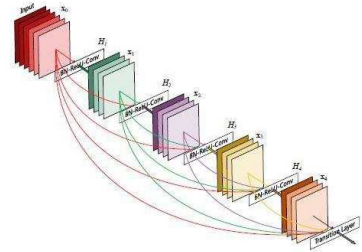}}
  \caption{Disease classification}
\end{figure}
%
%

Because of the serious data imbalance problem in this task, we select 2 directions for disease classification. In the first direction, a hierarchical structure is adopted to make even the number of training samples from different classes in each level, as shown in Fig.\ref{hierarchical}. The diseases are divided into 3 levels: in level 1, we need to classify NV and other diseases; in level 2, we need to classify MEL, BKL and other diseases; in level 3, we need to classify BCC, AKIEC, DF and VASC. In the other direction, we make even the training samples from different classes by using data augmentation. Different augmentation strategies are adopted for different classes and for the augmented training dataset, we have 20000 samples for each class. Both methods are based on the DenseNet-169 model, as shown in Fig.\ref{densenet}.

In the training stage, the parameters of the DenseNet-169 model can be summarized as follows: the backbone network is DenseNet, the training image resolution is $256 \times 340$ and the learning rate is set to $0.1$ with Adam Optimizer. In validation dataset, for the method using the hierarchical structure, the balanced classification performance is $0.621$; and for the method using direct training data augmentation, the balanced classification performance is $0.751$.

\section*{References}

[1] Hengshuang Zhao, Jianping Shi, Xiaojuan Qi, Xiaogang Wang \ \& Jiaya Jia\ (2017)
Pyramid Scene Parsing Network. {\it CVPR}, pp.\ 6230-6239

[2] Gao Huang, Zhuang Liu, Laurens van der Maaten \ \& Kilian Q. Weinberger\ (2017)
Densely Connected Convolutional Networks. {\it CVPR}, pp.\ 2261-2269.

[3] Patrick Ferdinand Christ, Mohamed Ezzeldin A. Elshaer, Florian Ettlinger, Sunil Tatavarty, Marc Bickel, Patrick Bilic, Markus Remp
er, Marco Armbruster, Felix Hofmann, Melvin D'Anastasi, Wieland H. Sommer, Seyed-Ahmad Ahmadi \ \& Bjoern H. Menze\ (2016)
Automatic Liver and Lesion Segmentation in CT Using Cascaded Fully Convolutional Neural Networks and 3D Conditional Random Fields {\it MICCAI}, pp.\ 415-423

[4] R Gonzalez \ \& R Woods\ (2010) {\it Digital Image Processing}, Prentice Hall International

[5] GD Finlayson\ (2004) Shades of gray and colour constancy. {\it Twelfth Color Imaging Conference}, pp.\ 37-41

\end{document}